\documentclass[10pt,conference,a4paper]{IEEEtran}

\usepackage{amsmath}
\usepackage{amssymb}
\usepackage[ruled, linesnumbered, vlined]{algorithm2e}
\usepackage{graphicx}
\usepackage{multirow}
\usepackage{subfigure}
\usepackage{multicol}
\usepackage{threeparttable}
\usepackage{booktabs}
\usepackage{color}

\newtheorem{ETA learning}{Definition}[section]
\newcommand{\R}{\textsl{RNML-ETA}\xspace}
\newcommand{\vct}[1]{\boldsymbol{#1}} 
\newcommand{\mat}[1]{\boldsymbol{#1}} 

\ifCLASSINFOpdf
\else
\fi
\begin{document}
%
\title{Road Network Metric Learning for\\ Estimated Time of Arrival}

\author{\IEEEauthorblockN{Yiwen Sun\IEEEauthorrefmark{1},
Kun Fu\IEEEauthorrefmark{2},
Zheng Wang\IEEEauthorrefmark{2},
Changshui Zhang\IEEEauthorrefmark{1} and
Jieping Ye\IEEEauthorrefmark{2}}
\IEEEauthorblockA{\IEEEauthorrefmark{1}Department of Automation, Tsinghua University, \\
State Key Lab of Intelligent Technologies
and Systems, \\
Institute for Artificial Intelligence, Tsinghua University (THUAI), \\ 
Beijing National Research Center for Information Science and Technology (BNRist), Beijing, China\\
Email: syw17@mails.tsinghua.edu.cn, zcs@mail.tsinghua.edu.cn}
\IEEEauthorblockA{\IEEEauthorrefmark{2}DiDi AI Labs, Beijing, China\\
Email: \{fukunkunfu, wangzhengzwang, yejieping\}@didiglobal.com}}



\maketitle

\begin{abstract}
Recently, deep learning have achieved promising results in Estimated Time of Arrival (ETA), which is considered as predicting the travel time from the origin to the destination along a given path. One of the key techniques is to use embedding vectors to represent the elements of road network, such as the links (road segments). However, the embedding suffers from the data sparsity problem that many links in the road network are traversed by too few floating cars even in large ride-hailing platforms like Uber and DiDi. Insufficient data makes the embedding vectors in an under-fitting status, which undermines the accuracy of ETA prediction. To address the data sparsity problem, we propose the Road Network Metric Learning framework for ETA (\R). It consists of two components: (1) a main regression task to predict the travel time, and (2) an auxiliary metric learning task to improve the quality of link embedding vectors. We further propose the triangle loss, a novel loss function to improve the efficiency of metric learning. We validated the effectiveness of \R on large scale real-world datasets, by showing that our method outperforms the state-of-the-art model and the promotion concentrates on the cold links with few data.
\end{abstract}


%
\IEEEpeerreviewmaketitle

\section{Introduction}
\label{sec:Introduction}
Intelligent Transportation System (ITS) aims to explore better transportation options for human beings and better relationships among users, vehicles and transportation infrastructures \cite{dimitrakopoulos2010intelligent,figueiredo2001towards}. Nowadays, with massive spatio-temporal data, artificial intelligence plays more and more important role in ITS by leveraging data-driven methods to analyze the traffic patterns, and has obtained promising results in many tasks of ITS \cite{zhang2011data,wang2018learning,guo2019attention}.

\begin{figure}[t]
    \centering
\includegraphics[width=0.95\linewidth]{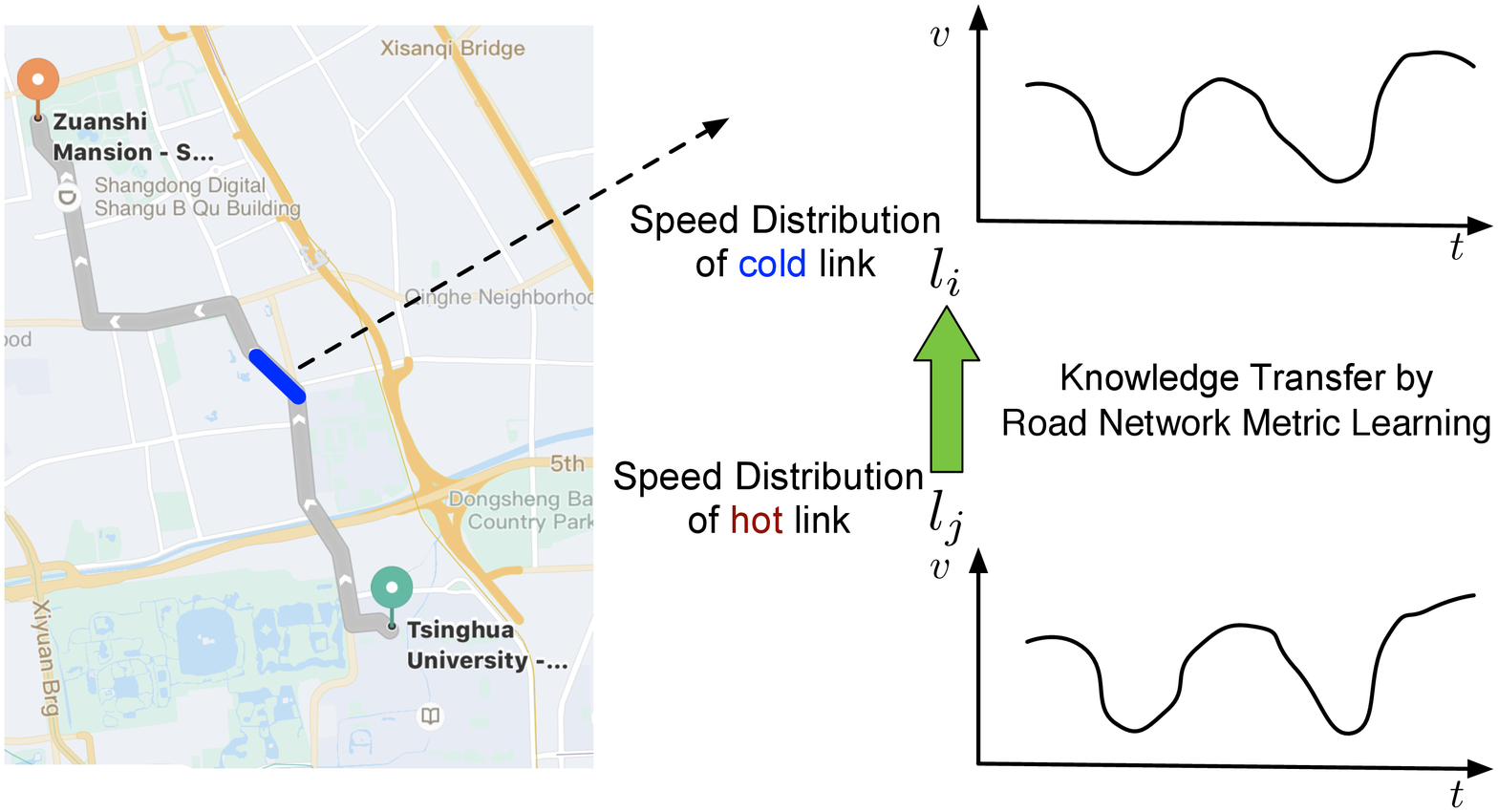}
\caption{The conceptual demonstration of \R.  The left part shows a real case in which the ETA system predicts the travel time along the route starting from the greed pin to the red pin. The route consists of a sequence of links. To alleviate the data sparsity problem, we propose to transfer the knowledge of hot links to the cold links by metric learning. The links' similarity is measured using their speed distrubtion.}
\label{fig:sketch}
\end{figure}

Estimated Time of Arrival (ETA) is one of the most fundamental and challenging problems in ITS. It is considered as predicting the travel time from an origin location to a destination location along a given route. An ETA model enables the transportation system to efficiently schedule the vehicles to control the increasing urban traffic congestion \cite{ccolak2016understanding}. Due to the rapid growth of ride-hailing apps such as Uber and DiDi, ETA has attracted more and more attention in recent years. An accurate ETA system can significantly improve the operating efficiency of the ride-hailing platforms by influencing  route planning, navigation, carpooling, vehicle dispatching and scheduling. The left part of Fig.~\ref{fig:sketch} show a real case of ETA.

Existing ETA methods can be divided into two categories. The fist one is the additive methods that explicitly predict the travel time for each road segment and give the total travel time of a route by assembling the ingredients' travel time. These methods have intuitive interpretability, but the prediction may be inaccurate when local errors are accumulated. The other one is the overall methods that directly predict the overall travel time of the route, by formulating ETA as a regression problem. For example, the Wide-Deep-Recurrent model (WDR) \cite{wang2018learning} takes neural network to predict the travel time based on a rich set of input features. This kind of methods avoid the local error accumulation but have relatively weak interpretability because of using black-box model.

We refer to the road segments as links in the remaining part of this paper. The technique of embedding \cite{bengio2003neural,mesnil2013investigation,mikolov2013distributed} is widely used, especially in deep learning ETA models, to capture the spatio-temporal patterns of link as it is one of the most fundamental element in the road network. Each link is represented by an embedding vector which encodes the link's semantic information through sufficient iterations during the training process. Though the ride-hailing platforms collect millions of trajectories per day, the embedding vectors still suffers from the data sparsity problem of road network that many links are traversed by too few floating cars. For cold links, which are covered by few trajectories, the training of their embedding vectors may end in an under-fitting status. Thus, the travel time estimation may have large error if a route goes through cold links.

To alleviate the data sparsity problem, we propose a novel ETA model named as \R. The model leverages multi-task learning \cite{caruana1997multitask} and consists of a main task predicting the travel time and an auxiliary task performing the metric learning, in which the similarity between links are measured by their speed distribution. Via the metric learning, similar links get close and dissimilar links get far away in the embedded space. Thus, the embedding vectors of cold links get sufficient training, which significantly improves the ETA accuracy. Moreover, we propose a novel loss function, the triangle loss, for metric learning to take more interaction into consideration in on update. To achieve this, we switch the roles of links among the anchor, positive and negative samples. A conceptual demonstration of \R is given in Fig.~\ref{fig:sketch}. 

The main contributions of this paper are three-fold:

\begin{itemize}
    
    \item To our best knowledge, \R is the first deep learning method that effectively addresses the  data sparsity problem of road network.

    \item We propose a novel metric learning framework to improve the quality of link embedding vectors. The similarity of links can be measured using the speed distribution of links which can be computed from existing ETA data, requiring no extra information. We also propose the novel triangle loss to improve the learning efficiency of metric learning.
   
    \item We conducted comprehensive evaluation of our method on large scale real-world datasets containing over 100 million trajectories. The experimental results validated that \R significantly improves the performance compared to a state-of-the-art deep learning method.
    
\end{itemize}

The rest of this paper is organized as follows. Section~\ref{sec:RELATED WORK} reviews the related works. Section~\ref{sec:METHODOLOGY} introduces our method \R in detail. Section~\ref{sec:EXPERIMENT} gives the experimental results on the large-scale real-world datasets. Section~\ref{sec:CONCLUSION} is a conclusion of this paper.

\section{RELATED WORK}
\label{sec:RELATED WORK}
\textbf{Estimated Time of Arrival.} 
As one of the fundamental problems in intelligent transportation system, ETA attracts an extensive study in both academic and industrial communities. ETA models can be divided into two categories. The first category is the additive methods that explicitly estimate the travel time for each link and give the prediction of a route by assembling the ingredients' travel time. Rule-based method can be used in the estimation of link travel time. For example, a simple rule dividing the link length by the link travel speed is widely used in the industry. Learning-based methods, such as the dynamic bayesian network \cite{hofleitner2012learning}, gradient boosted regression tree~\cite{zhang2016urban}, least-square minimization~\cite{zhan2013urban} and pattern matching~\cite{chen2013dynamic} are also used to mine the traffic patterns and predict the link's travel time. The data sparsity problem of road network is discussed in \cite{wang2014travel} that a part of links are traversed by too few trajectories. To alleviate the data sparseness , the authors of  \cite{wang2014travel} propose to represent the trips as a tensor and utilize tensor decomposition to complete the missing values. However, dealing with data sparsity is still a challenging problem for ETA.

The second category is the overall methods that directly predict the overall travel time of the given route. Early methods such as TEMP~\cite{wang2016simple} and time-dependent landmark graph~\cite{yuan2011t} use traditional machine learning methods to predict the travel time. 
Recently, due to the bloom of deep learning \cite{lecun2015deep,krizhevsky2012imagenet,larochelle2009exploring}, neural network models for ETA are in a rapid development. MURAT~\cite{li2018multi} uses feed-forward neural networks to predict the travel time from the origin to the destination without a given path. Multi-task learning and graph embedding are used in MURAT to narrow the accuracy gap to the path-based methods. DeepTTE~\cite{wang2018will} proposes a geo-convolution operation to encode the coordinate information and uses recurrent neural network to learn the travel time along a GPS sequence. Since GPS sequence cannot be acquired until the trip is finished, DeepTTE resamples the GPS points by uniform distance at training stage and generates pseudo points according to a planned route at inference stage. WDR model~\cite{wang2018learning} uses a wide linear part and a deep neural network to learn the trip-level information, and a recurrent neural network to learn the fine-grained sequential information in the route. 
The authors of~\cite{fu2019deepist,lan2019travel} transform the map information into the image sequence, and adopt convolutional neural network to mine spatial correlations for ETA.
In these deep learning methods, the embedding of geographical elements, such as the link embedding in \cite{li2018multi,wang2018learning} and the grid embedding in \cite{zhang2018deeptravel}, plays an important role. The embedding technique suffers from the data sparsity problem as well, because insufficient data makes the embedding vectors in an under-fitting status.

\textbf{Metric learning.} 
The goal of metric learning is to learn a representation function that maps objects into an embedded space. The distance in the embedded space should preserve the objects' similarity --- similar objects get close and dissimilar objects get far away. Various loss functions have been developed for metric learning. For example, the contrastive loss \cite{chopra2005learning} guides the objects from the same class to be mapped to the same point and those from different classes to be mapped to different points whose distances are larger than a margin. Triplet loss \cite{schroff2015facenet} is also popular, which requires the distance between the anchor sample and the positive sample to be smaller than the distance between the anchor sample and the negative sample. The case with one positive sample and multiple negative samples is extended in \cite{sohn2016improved}. Metric learning often suffers from slow convergence, partially because the loss only captures limited interaction in one update.
\section{METHODOLOGY}
\label{sec:METHODOLOGY}

We describe the road network as a set of links $\{l = 1, 2, \cdots, M\}$, where $M$ is the total link number in the map and $l$ is the link ID ranging from 1 to $M$.  We then give the definition of ETA learning problem which is essentially a regression task:

\begin{ETA learning}
\textbf{ETA Learning}. Suppose we have a collection of historical trips $\{s_i, e_i, d_i, \vct{p}_i\}_{i=1}^N$, where $N$ stands for the total trip number, $s_i$ is the departure time, $e_i$ is the arriving time, $d_i$ is the driver ID and $\vct{p}_i$ is the travel path for $i$-th trip. Our goal is to fit a model that can predict the travel time estimation $y'_i$ given the departure time, the driver ID and the travel path. The ground-truth travel time $y_i$ can be computed as $y_i = e_i - s_i$. The travel path $\vct{p}_i$ is represented as a sequence of links $\vct{p}_i = \{l_{i1}, l_{i2}, \cdots, l_{iT_i}\}$, where $l_{ij}$ is the ID of $j$-th link in the $i$-th sequence and $T_i$ is the sequence length of $\vct{p}_i$.
\end{ETA learning}


We introduce the overall framework of the proposed method in  Section \ref{subsec:RnmlEta}, define the measurement of link similarity in Section \ref{subsec:LinkSimilarity} and introduce the details of our metric learning loss in Section \ref{subsec:TriangleLoss}.

\subsection{Overall Framework}
\label{subsec:RnmlEta} 
We first construct a rich feature set from the raw information of trips. For example, according to the departure time, we can obtain the time slice in a day (every 5 minutes) and the day of week. The features can be categorized into two types: (1) the sequential features which are extracted from the travel path $\vct{p}_i$. For a link $l_{ij}$, we denote its feature vector as $\vct{x}_{ij}$, and get a feature matrix $\mat{X}_i = [\vct{x}_1, \cdots, \vct{x}_{T_i}]$ for the $i$-th trip. Note that the sequential feature has variable size --- in other words,  the column number of $\mat{X}_i$ is decided by the path length; and (2) the non-sequential features which are irrelative to the travel path, e.g day of the week. They are represented as a feature vector $\vct{z}_i$ with fixed size.

The link embedding vector is an important component of  the link feature vector $\vct{x}_{ij}$. For link with ID=$l_{ij}$, we look up an embedding table $\mat{E}_L\in \mathbb{R}^{20 \times M}$, and use its $l_{ij}$-th column $\mat{E}_L(:, l_{ij})$ as a distributional representation for the link~\cite{bengio2003neural} . The $\mat{E}_L$ is randomly initialized and will be updated in the training process by gradient descending to encode semantic information of links. The link feature vector is a concatenation of  $\mat{E}_L(:, l_{ij})$, the link length $len(l_{ij})$ and the link's travel speed $v_{ij}$:
\begin{equation}
\vct{x}_{ij} = [\mat{E}_L(:, l_{ij}); len(l_{ij}); v_{ij}].
\end{equation}
The link's length is obtained by geographical survey and the travel speed is the average speed of the floating cars that traversed the link within the latest time window (e.g 10 minutes).

Data amount significantly affects the quality of embedding vectors. For example in the natural language processing field, Word2vec~\cite{mikolov2013distributed} cannot generate meaningful embedding vectors for rare words that occur in very limited sentences. In ride-hailing platforms, the data coverage on road network is still not satisfactory though there are already millions of floating cars. A part of links are traversed by only a few or even zero trajectories. We refer to those traversed by plenty of trips as hot links, and those traversed by only a few or even zero trips as cold links. The hot links' embedding vectors can be well trained with sufficient iteration. However, the training of cold links' embedding vectors is often ended in an under-fitting status, which undermines the accuracy of ETA prediction. 

To improve the embedding quality of cold links, we propose the Road Network Metric Learning ETA (\R), whose training process consists of two tasks. The main task is to predict the travel time, while the auxiliary task is to regularize the link embedding vectors by transferring the knowledge of road network patterns from hot links to cold links. The metric learning in the auxiliary task can help to place the embedding vector of a cold link in a proper position in the embedded space, by reducing the distance to its similar hot links. The loss function of \R is:
\begin{equation}
L = (1 - \beta) \cdot L_{main} + \beta \cdot L_{aux},
\end{equation}
where $\beta$ is a hyper-parameter to balance the trade-off between the main task and the auxiliary task.

We choose Wide-Deep-Recurrent (WDR) model \cite{wang2018learning}, a state-of-the-art ETA model, to accomplish the main task. The three components of WDR model includes: (1) a \emph{wide} module memorizing the historical patterns in data by constructing a second order cross product and an affine transformation of the non-sequential feature $\vct{z}_i$; (2) a \emph{deep} module improving the generalization ability by feeding $\vct{z}_i$ into a Multi-Layer Perceptron (MLP), which is a stack of fully-connected layers with ReLU \cite{krizhevsky2012imagenet} activation functions; and (3) a \emph{recurrent} module providing a fine-grained modeling on the sequential feature $\mat{X}_i$ via Long-Short Term Memory network (LSTM) \cite{hochreiter1997long}, which can capture the spatial and temporal dependency between links.

We denote the outputs of the wide module as $\vct{h}_i^{(w)}$, the output of the deep module as $\vct{h}_i^{(d)}$, and the last hidden state of LSTM as $\vct{h}_i^{(T_i)}$. The travel time prediction is given by a regressor, which is also a MLP, based on the concatenation of the outputs:
\begin{equation}
y_i' = MLP(\vct{h}_i^{(w)}, \vct{h}_i^{(d)}, \vct{h}_i^{(T_i)}).
\end{equation}

\begin{figure}[t]
    \centering
\includegraphics[width=0.99\linewidth]{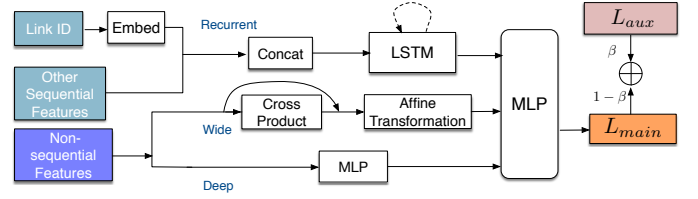}
\caption{The overall architecture of \R. The loss function consists of two aspects: (1) the main task uses a Wide-Deep-Recurrent model to learn the travel time prediction, and (2) the auxiliary task uses metric learning to improve the quality of link embedding vectors.}
\label{fig:arch}
\end{figure}

The hidden state sizes in the deep module, the LSTM and the regressor MLP are all set to 128. The hidden state and memory cell of LSTM are initialized as zeros. We choose Mean Absolute Percentage Error (MAPE) as the loss function of the main task:
\begin{equation}
L_{main} = {\frac{\hbox{1}}{N}} \sum_{i = 1}^{N}{\frac{\left\vert y_{i} - y_{i}' \right\vert}{y_{i}}} , 
\label{eq:MAPE}
\end{equation}
where $y_i$ is the ground-truth travel time. The overall architecture of \R and the main task workflow are visualized in Fig.~\ref{fig:arch}. The details of the auxiliary task will be introduced in the following sections.

\subsection{Link Similarity}
\label{subsec:LinkSimilarity}
To apply metric learning on the link embedding vectors, a similarity measurement of links should be defined. Since the link's travel speed essentially reflects how long a car is expected to take to pass through the link, the speed distribution across different time could be used to depict the traffic characteristic of the link. We construct a series of time bins $\{\tau_1, \tau_2, \cdots, \tau_K \}$ for a day. These time bins are ensured to be non-overlapped: $\tau_i \cap \tau_j = \emptyset, \forall i \neq j$; and their union covers the whole day: $\tau_1 \cup \tau_2 \cup \cdots \cup \tau_K = 24h$. We then statistic the average travel speed for link $l$ and time bin $\tau_k$ by computing:
\begin{equation}
\begin{aligned}
\bar{v}_k(l) &= \frac{1}{Z}  \sum_{i=1}^N \sum_{j=1}^{T_i} v_{ij} I_{s_i \in \tau_k}  I_{l_{ij} = l}, \\
Z &=  \sum_{i=1}^N \sum_{j=1}^{T_i} I_{s_i \in \tau_k} I_{l_{ij} = l},
\end{aligned}
\end{equation}
where $v_{ij}$ is the travel speed feature of $j$-th link in $i$-th trip, and $I_{cond}$ is an indicator that $I_{cond}=1$ if $cond$ is satisfied and $I_{cond}=0$ otherwise.
Intuitively, we find a subset of the link $l$'s travel speed features by selecting those whose departure time belongs to the time bin $\tau_k$, and then compute the average on the subset. In practice, we use a configuration of $K=3$ time bins with $\tau_1$ from 5 a.m to 11 a.m representing the morning peak, $\tau_2$ from 4 p.m to 10 p.m representing the evening peak and $\tau_3$ taking the remaining hours representing the off-peak time.

We further scale the speeds to be within $[0, 1]$ by applying $\widetilde{v}_k(l) = (v_k(l) - a) / (b - a)$, where $a$ and $b$ are the minimum and maximum of  $\{ v_k(l), k=1 \cdots K, l = 1 \cdots M \}$. We finally get a normalized speed histogram of link $l$:
\begin{equation}
\widetilde{\vct{v}}(l) = [\widetilde{v}_1(l),~ \widetilde{v}_2(l),~ \widetilde{v}_3(l)]^T.
\end{equation}

A difference matrix $\mat{Q} \in \mathbb{R}^{M \times M}$ can be computed as follows:
\begin{equation}
{Q}_{ij} = {Q}_{ji} = \Vert \widetilde{\vct{v}}(i)- \widetilde{\vct{v}}(j) \Vert_2,
\end{equation}
where $Q_{ij}$ is the element of $\mat{Q}$ measuring the difference between links with ID=$i$ and ID=$j$. Smaller difference means larger similarity. The similarity based on speed histogram shows advantages on two aspects. Firstly, the ETA is mostly determined by the traffic condition and is partially influenced by personalized factors such as the driving habit. The latest average speed is a good reflection of the traffic condition. If two links have similar speed distribution, they should also have similar impact on the ETA prediction. Secondly, the speed histogram does not rely any extra information and can be computed directly from the data used in the main task, which facilitates the method implementation.

\subsection{Triangle Loss}
\label{subsec:TriangleLoss}
Links with similar characteristic are expected to be closer in the embedded space and those with dissimilar characteristic are expected to be farther.
With this end in view, we propose a novel metric learning loss function, named as triangle loss. Suppose we have three links with ID=$l_i, l_j, l_k$ and the corresponding differences $Q_{l_i l_j}$, $Q_{l_j l_k}$ and $Q_{l_i l_k}$, without loss of generality, we assume:
\begin{equation}
Q_{l_i l_j} < Q_{l_j l_k} < Q_{l_i l_k}.
\label{eq:relationship}
\end{equation}

\begin{figure}[t]
    \centering
\includegraphics[width=0.6\linewidth]{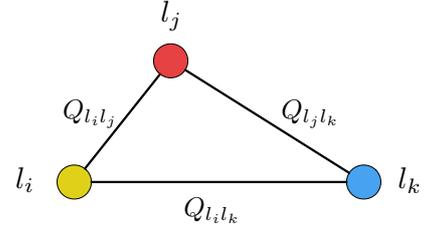}
\caption{The distances forms a triangle and the order of their edge lengths should satisfy the relation in Eq. \ref{eq:relationship}.}
\label{fig:triangle loss}
\end{figure}

We then compute the Euclidean distances between the embedding vectors of link $l_i$, $l_j$ and $l_k$. For example:
\begin{equation}
D_{l_i l_j}= \Vert \widetilde{\mat{E}}_L(:, l_i) - \widetilde{\mat{E}}_L(:, l_j) \Vert_{2},
\end{equation}
where $\widetilde{\mat{E}}_L(:, l_i) = \mat{E}_L(:, l_i) / \Vert \mat{E}_L(:, l_i) \Vert_2$ is the L-2 normalized embedding vector. The three distances $D_{l_i l_j}$, $D_{l_j l_k}$ and $D_{l_i l_k}$ forms a triangle. We aims to restrict the lengths of the triangle edges to be in the same order as in Eq. \ref{eq:relationship}, which derives three inequations:
\begin{equation}
\begin{aligned}
D_{l_i l_j}^2 + \alpha_1 < D_{l_j l_k}^2, \\ 
D_{l_i l_j}^2 + \alpha_2 < D_{l_i l_k}^2, \\
D_{l_j l_k}^2 + \alpha_3 < D_{l_i l_k}^2
\end{aligned}
\end{equation}
where $\alpha_{1}$, $\alpha_{2}$ and $\alpha_{3}$ are required margins. Unlike the triplet loss~\cite{schroff2015facenet} which has only one restriction that the distance between anchor and positive sample should be smaller than the distance between anchor and negative sample, the links in our method take turns to act as the anchor. This enables a more efficient metric learning in one update and thus accelerates the convergence. Fig.~\ref{fig:triangle loss} gives a visualized demonstration. The triangle loss is in the form of:
\begin{equation}
\begin{split}
L_{aux}= \frac{1}{U} \sum_{l_i, l_j, l_k} \bigg( & \gamma_1 \Big[D_{l_i l_j}^{2}-D_{l_j l_k}^{2} + \alpha_{1}\Big]_{+} \\ 
& + \gamma_2 \Big[D_{l_i l_j}^{2}-D_{l_i l_k}^{2} + \alpha_{2}\Big]_{+} \\
& + \gamma_3 \Big[D_{l_j l_k}^{2}-D_{l_i l_k}^{2} + \alpha_{3}\Big]_{+} \bigg),
\end{split}
\end{equation}
where the operator $[x]_+ = max(x, 0)$ and $U$ is the number of possible triangles in the training set, $\gamma_{1}$, $\gamma_{2}$ and $\gamma_{3}$ are hyper-parameters to adjust the weights of the three distances. The auxiliary task and main task are simultaneously optimized via gradient descending. For a mini-batch of trips, we first compute the loss of the main task, and then compute the auxiliary loss by randomly combining triangles with all the links in the trips.

\section{EXPERIMENT}
\label{sec:EXPERIMENT}

The evaluation is on large scale real-world datasets collected in DiDi platform. We will introduce the datasets, the competing methods, the implementation details and the experimental results in sequence.

\subsection{Dataset}
We collected massive floating car trajectories of Beijing in 2018 in DiDi platform. The trajectories are split into \emph{pickup} and \emph{trip} datasets according to the driver's working status. A \emph{pickup} trajectory starts when a driver responds to a passenger's request and ends when he/she picks up the passenger. A \emph{trip} trajectory starts when the passenger gets on board and ends when arriving the destination. For each dataset, we use 25 weeks of data as training set and the following 2 weeks as validation set and test set, respectively. We remove the outliers with extremely short travel time ($<$60s) and extremely high average speed ($>$120km/h). The data statistics are summarized in Table~\ref{tbl:dataset}.

\begin{table}[htb]
\centering
    \caption{Statistics of datasets}
\label{tbl:dataset}
\begin{tabular}{c c c c} 
 \toprule
 &  size  & \emph{pickup} & \emph{trip} \\
 \midrule
 training set    &   25 weeks  &   111.0M      & 105.5M  \\
 validation set  &   1 week   &   4.0M        &  4.5M   \\
 test set        &   1 week   &   4.1M        &  3.9M   \\
\# traversed link  &     -       &   1.2M      &  1.3M \\
\bottomrule
\end{tabular}
\end{table}

\begin{figure}[htb]
    \centering
\includegraphics[width=0.95\linewidth]{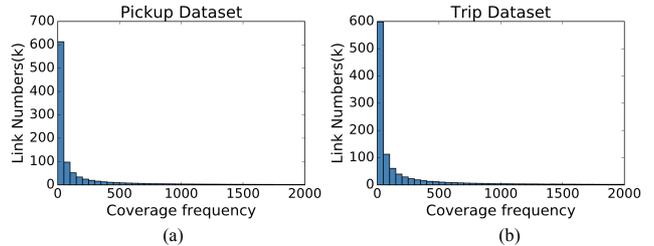}
\caption{Statistics of link coverage frequency. For both \emph{pickup} and \emph{trip} datasets, the links concentrate on the bands with small number of traversing trajectories.}
\label{fig:number}
\end{figure}

The links are from a wide range of roads, such as private community roads, local streets and urban freeways. As shown in Table \ref{tbl:dataset}, the \emph{trip} dataset covers more links than the \emph{pickup} dataset. However, both the datasets suffer from the road network sparsity problem that most of the links are short of data. To demonstrate it, we plot the histogram of link coverage frequency in Fig.~\ref{fig:number}. Though with over 0.1 billion of trajectories, there is a significant number of cold links that are traversed by only a few times in about half a year (25 weeks). The median coverage frequencies of link are 42 on \emph{pickup} and 69 on \emph{trip}.

\subsection{Competing Methods}
We compare the proposed \R with the following competitors.

(1) Route-ETA: a representative method in industrial application. In this solution, the travel time estimation for each link is made by dividing the link length by the link travel speed. The waiting time at each intersection is mined from the historical data. Given a route, the total travel time is predicted as the sum of each link's travel time and each intersection's waiting time. Route-ETA has very fast inference speed but its accuracy is often far from satisfactory compared to deep learning methods. 

(2) WDR~\cite{wang2018learning}: a deep learning method achieving the state-of-the-art performance in ETA problem. Since it is the model used in our main task, the comparison between WDR and \R evaluates the benefit of the auxiliary task.

(3) WDR-no-link-emb: a variant of WDR that removes the link embedding technique. The main purpose of using this model is to quantify the contribution of link embedding vectors, of which the \R is aiming to improve the quality.

Besides the Mean Absolute Percentage Error (MAPE), which is used as objective function in the main task, we also take Mean Absolute Error (MAE) and Root Mean Square Error (RMSE) as the evaluation metrics. The computations are:
\begin{equation}
\begin{aligned}
\text{MAE} &= \frac{1}{N} \sum_{i = 1}^{N}{ \left\vert y_{i} - y_{i}' \right\vert} , \\
\text{RMSE} &= \left[ \frac{1}{N} \sum_{i = 1}^{N}{ \left( y_{i} - y_{i}' \right)^2} \right]^{1/2}.
\end{aligned}
\end{equation}

\subsection{Implementation Details}
\label{sec:implementation}
The neural networks in WDR, WDR-no-link-emb and \R are implemented in PyTorch~\cite{paszke2019pytorch}, and the training is accelerated on a single NVIDIA P40 GPU. We use a mini-batch size of 256 and set the maximal iteration number to 7 millions.
The hyper-parameters of \R are selected by the results on validation set. We use margins $\alpha_1 = \alpha_3 = 0.005$, $\alpha_2 = 0.02$ and weights $\gamma_1 = \gamma_3 = 0.3$, $\gamma_2=0.4$ in the triangle loss for both \emph{pickup} and \emph{trip} datasets. The task weight $\beta$ is 0.52 for \emph{pickup} and 0.35 for \emph{trip}. All the parameters, such as the MLP weights and the embedding vectors, are jointly trained using Adam \cite{kingma2015adam} optimizer, which is a stochastic gradient descending method. Adam can adaptively adjust the step size according to the historical gradients and thus accelerate the convergence. The learning rate is set to 0.0002.

\subsection{Experimental Results}
We list the results of \emph{pickup} data in Table~\ref{tbl:pickup} and \emph{trip} data in Table~\ref{tbl:trip}, and mark the best scores by bold font. The proposed method \R outperforms all the competitors on both datasets. The metric learning component significantly improves the main task model's accuracy to predict the travel time. For example, \R reduces $2.62\%$ RMSE on \emph{pickup} data and reduces $1.19\%$ MAPE on \emph{trip} data compared to WDR. The importance of link embedding technique is also validated that it brings $7.0\%$ and $7.9\%$ reduction on MAPE for \emph{pickup} and \emph{trip} data, respectively (WDR-no-link-emb v.s. WDR). Moreover, it can be observed that there is a large performance gap between the simple rule-based model Route-ETA and the deep learning models.

\begin{table}[htb]
\centering
    \caption{Results of the pickup dataset}
\label{tbl:pickup}
\begin{tabular}{c c c c} 
 \toprule
 &  MAPE (\%)  & MAE (sec) & RMSE (sec) \\
 \midrule
 Route-ETA  & $25.010$ & 69.008 & 106.966  \\
 WDR-no-link-emb & $20.845$ & 59.018 & 95.876  \\
 WDR & $19.386$ & 54.686 & 89.976  \\
 \R & $\textbf{19.215}$ & \textbf{53.546} & \textbf{87.617} \\
\bottomrule
\end{tabular}
\end{table}

\begin{table}[htb]
\centering
    \caption{Results of the trip dataset}
\label{tbl:trip}
\begin{tabular}{c c c c} 
 \toprule
 &  MAPE(\%)   & MAE (sec) & RMSE (sec) \\
 \midrule
 Route-ETA & $15.440$ & 150.560 & 248.736 \\
 WDR-no-link-emb & $12.742$ & 117.337  & 197.652 \\
 WDR & $11.737$ & 108.919 & 186.083  \\
 \R & $\textbf{11.597}$ & \textbf{108.519} & \textbf{185.897} \\
\bottomrule
\end{tabular}
\end{table}

\begin{figure}[tb]
    \centering
\includegraphics[width=0.99\linewidth]{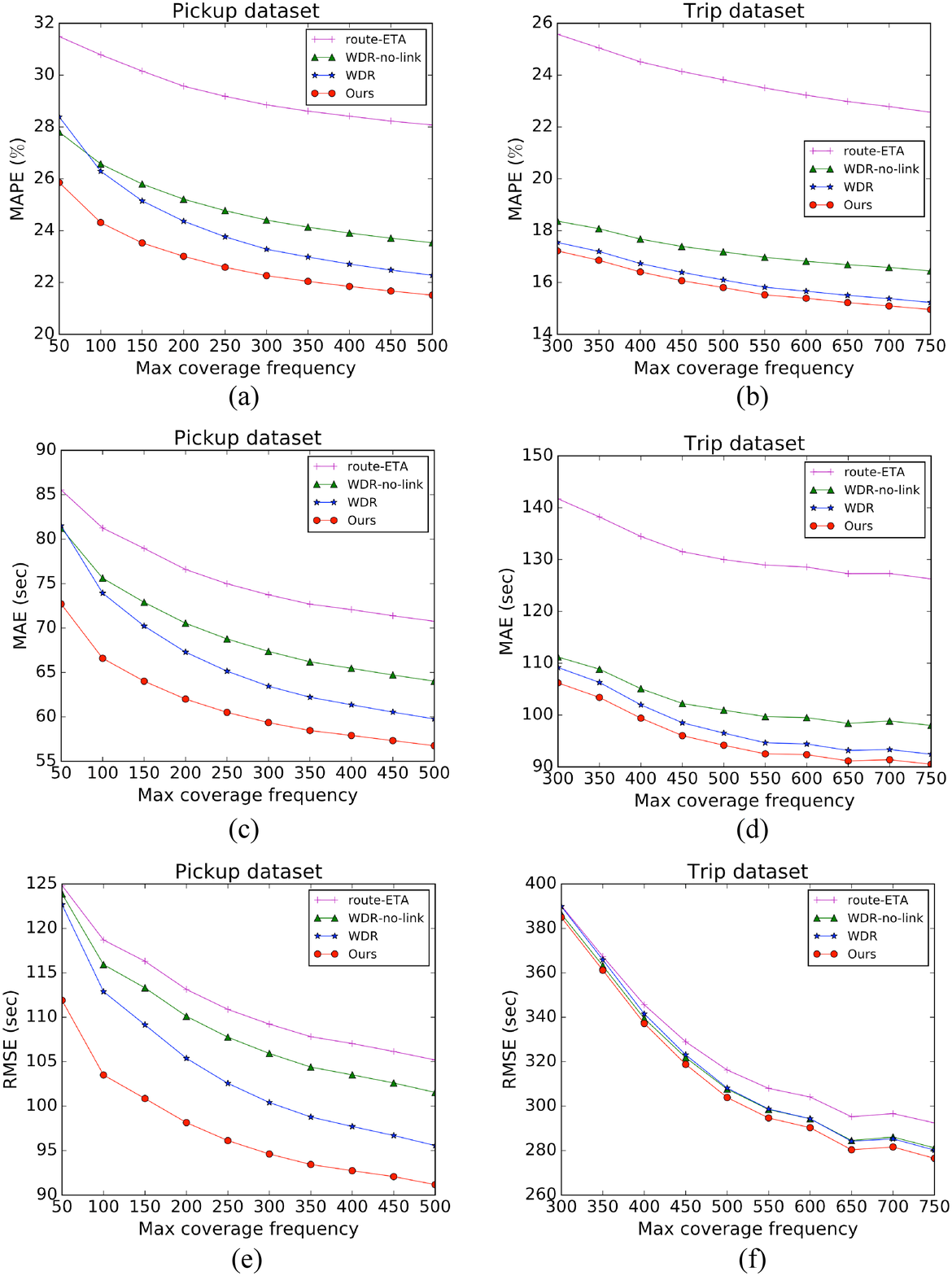}
\caption{Results of the finer evaluation on subsets with different link coverage level. For a threshold $\delta$, we keep the trajectory that at least $25\%$ of the contained links have coverage frequencies less than $\delta$. The 6 subfigures stand for (a) MAPE on \emph{pickup} data, (b) MAPE on \emph{trip} data, (c) MAE on \emph{pickup} data, (d) MAE on \emph{trip} data, (e) RMSE on \emph{pickup} data and (f) RMSE on \emph{trip} data.}
\label{fig:sparse}
\end{figure}

The results in Table~\ref{tbl:pickup} and Table~\ref{tbl:trip} show the overall accuracy on all the links. Since \R mainly aims to improve the embedding quality of cold links, its contribution needs a finer evaluation which reports the metrics at different link coverage level. Thus, we select a series of subsets from the dataset by restricting the link coverage frequency in the trajectory. Specifically, we keep a trajectory if at least $25\%$ of the contained links have coverage frequencies less than a threshold $\delta$, and drop the trajectory otherwise. By varying $\delta$ from 50 to 500 on \emph{pickup} data and from 300 to 750 on \emph{trip} data in a step of 50, we obtain 10 subsets for each dataset. In subset with lower $\delta$, the trajectory contains more cold links. We then compute the metrics on these subsets and plot the curves in Fig.~\ref{fig:sparse}.

We take Fig.~\ref{fig:sparse} (a) as an example (the trends in other subfigures are similar).  As the threshold $\delta$ increases, the subset includes more hot links and the MAPE of WDR gradually decreases from $28\%$ to $23\%$, which is a large improvement for ETA problem. This phenomenon shows that links covered by more trajectories do have better prediction accuracy and supports the existence of the road network data sparsity problem. On the subset with $\delta=50$, our method \R outperforms WDR by more than 2 percentage in terms of MAPE. However, the gain on overall MAPE (Table~\ref{tbl:pickup}) is less than 0.2 percentage. Such a comparison validates the effectiveness of \R that it mainly improves the performance of cold links. As $\delta$ increases, \R achieves MAPE improvements up to $8.92\%$ on \emph{pickup} data and up to $1.99\%$ on \emph{trip} data.


\subsection{Influence of Hyper-parameter}
\begin{figure}[htb]
    \centering
\includegraphics[width=1.0\linewidth]{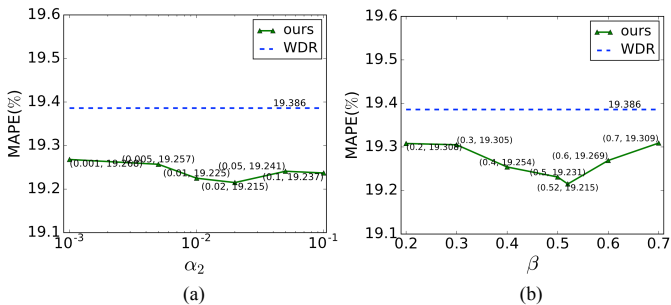}
\caption{The influence of hyper-parameters: (a) for the margin $\alpha_2$ in the triangle loss, and (b) for the weight balancing the main task and the auxiliary task. Though MAPE varies under different hyper-parameters, \R generally outperforms the competitor WDR, which demonstrates the robustness of our method.}
\label{fig:hyper}
\end{figure}

To explore the influence of hyper-parameters, we plot the performance curves of \emph{pickup} data in Fig.~\ref{fig:hyper} by varying the margin $\alpha_2$ and the task weight $\beta$, which are two representative hyper-parameters. The basic configuration is the same as in Section \ref{sec:implementation}, namely, $\alpha_1 = \alpha_3 = 0.005$, $\alpha_2 = 0.02$, $\gamma_1 = \gamma_3 = 0.3$, $\gamma_2=0.4$ and $\beta=0.52$.

The hyper-parameter $\alpha_2$ is a bit more special than $\alpha_1$ and $\alpha_3$, because it controls the gap between the longest edge and the shortest edge in the triangle loss. If this restriction is broken, it means that the model is far from our expected status and needs a stronger gradient to update the parameters. Usually, we set $\alpha_2 > \alpha_1 + \alpha_3$ and find that $0.02$ achieves the best performance according to the curve in Fig. \ref{fig:hyper} (a). Moreover, \R achieves better performance than WDR from $\alpha_2=0.001$ to $0.1$, which demonstrates that the superiority of \R is not sensitive to the margin hyper-parameter.

The task weight $\beta$ is to balance the trade-off between the main task and the auxiliary task. In extreme cases, \R degenerates to WDR if $\beta=0$ and degenerates to a pure metric learning model if $\beta=1$. Fig. \ref{fig:hyper} (b) shows that the advantage of \R over WDR is robust in a wide range of $\beta$ from $0.2$ to $0.7$ and that the best performance is achieved at $\beta=0.52$.

\section{Conclusion}
\label{sec:CONCLUSION}
In this paper, we propose a novel metric learning framework for ETA, named as \R, to address the data sparsity problem of road network. In the main task, we use WDR model to predict the travel time. In the auxiliary task, we first construct a difference matrix by computing the Euclidean distances between the links' speed distributions, and then use metric learning to get the similar links close and dissimilar links far away in the embedded space. The auxiliary task is aiming to improve the quality of embedding vectors of links. We conduct experiments on two large scale real-world datasets collected in DiDi platform. The results validated the effectiveness of \R by showing that it outperforms the state-of-the-art WDR model on all the evaluation metrics. A further experiment finely examines the gains for different types of link and find that \R significantly improves the accuracy for routes containing cold links.





%



\bibliographystyle{IEEEtran}
\bibliography{references} 

\begin{thebibliography}{10}
\providecommand{\url}[1]{#1}
\csname url@samestyle\endcsname
\providecommand{\newblock}{\relax}
\providecommand{\bibinfo}[2]{#2}
\providecommand{\BIBentrySTDinterwordspacing}{\spaceskip=0pt\relax}
\providecommand{\BIBentryALTinterwordstretchfactor}{4}
\providecommand{\BIBentryALTinterwordspacing}{\spaceskip=\fontdimen2\font plus
\BIBentryALTinterwordstretchfactor\fontdimen3\font minus
  \fontdimen4\font\relax}
\providecommand{\BIBforeignlanguage}[2]{{%
\expandafter\ifx\csname l@#1\endcsname\relax
\typeout{** WARNING: IEEEtran.bst: No hyphenation pattern has been}%
\typeout{** loaded for the language `#1'. Using the pattern for}%
\typeout{** the default language instead.}%
\else
\language=\csname l@#1\endcsname
\fi
#2}}
\providecommand{\BIBdecl}{\relax}
\BIBdecl

\bibitem{dimitrakopoulos2010intelligent}
G.~Dimitrakopoulos and P.~Demestichas, ``Intelligent transportation systems,''
  \emph{IEEE Vehicular Technology Magazine}, vol.~5, no.~1, pp. 77--84, 2010.

\bibitem{figueiredo2001towards}
L.~Figueiredo, I.~Jesus, J.~T. Machado, J.~R. Ferreira, and J.~M. De~Carvalho,
  ``Towards the development of intelligent transportation systems,'' in
  \emph{ITSC (Cat. No. 01TH8585)}.\hskip 1em plus 0.5em minus 0.4em\relax IEEE,
  2001, pp. 1206--1211.

\bibitem{zhang2011data}
J.~Zhang, F.-Y. Wang, K.~Wang, W.-H. Lin, X.~Xu, and C.~Chen, ``Data-driven
  intelligent transportation systems: A survey,'' \emph{IEEE Transactions on
  Intelligent Transportation Systems}, vol.~12, no.~4, pp. 1624--1639, 2011.

\bibitem{wang2018learning}
Z.~Wang, K.~Fu, and J.~Ye, ``Learning to estimate the travel time,'' in
  \emph{SIGKDD}.\hskip 1em plus 0.5em minus 0.4em\relax ACM, 2018, pp.
  858--866.

\bibitem{guo2019attention}
S.~Guo, Y.~Lin, N.~Feng, C.~Song, and H.~Wan, ``Attention based
  spatial-temporal graph convolutional networks for traffic flow forecasting,''
  in \emph{AAAI}, vol.~33, 2019, pp. 922--929.

\bibitem{ccolak2016understanding}
S.~{\c{C}}olak, A.~Lima, and M.~C. Gonz{\'a}lez, ``Understanding congested
  travel in urban areas,'' \emph{Nature communications}, vol.~7, no.~1, pp.
  1--8, 2016.

\bibitem{bengio2003neural}
Y.~Bengio, R.~Ducharme, P.~Vincent, and C.~Jauvin, ``A neural probabilistic
  language model,'' \emph{Journal of machine learning research}, vol.~3, no.
  Feb, pp. 1137--1155, 2003.

\bibitem{mesnil2013investigation}
G.~Mesnil, X.~He, L.~Deng, and Y.~Bengio, ``Investigation of
  recurrent-neural-network architectures and learning methods for spoken
  language understanding.'' in \emph{Interspeech}, 2013, pp. 3771--3775.

\bibitem{mikolov2013distributed}
T.~Mikolov, I.~Sutskever, K.~Chen, G.~S. Corrado, and J.~Dean, ``Distributed
  representations of words and phrases and their compositionality,'' in
  \emph{NeurIPS}, 2013, pp. 3111--3119.

\bibitem{caruana1997multitask}
R.~Caruana, ``Multitask learning,'' \emph{Machine learning}, vol.~28, no.~1,
  pp. 41--75, 1997.

\bibitem{hofleitner2012learning}
A.~Hofleitner, R.~Herring, P.~Abbeel, and A.~Bayen, ``Learning the dynamics of
  arterial traffic from probe data using a dynamic bayesian network,''
  \emph{IEEE Transactions on Intelligent Transportation Systems}, vol.~13,
  no.~4, pp. 1679--1693, 2012.

\bibitem{zhang2016urban}
F.~Zhang, X.~Zhu, T.~Hu, W.~Guo, C.~Chen, and L.~Liu, ``Urban link travel time
  prediction based on a gradient boosting method considering spatiotemporal
  correlations,'' \emph{ISPRS International Journal of Geo-Information},
  vol.~5, no.~11, p. 201, 2016.

\bibitem{zhan2013urban}
X.~Zhan, S.~Hasan, S.~V. Ukkusuri, and C.~Kamga, ``Urban link travel time
  estimation using large-scale taxi data with partial information,''
  \emph{Transportation Research Part C: Emerging Technologies}, vol.~33, pp.
  37--49, 2013.

\bibitem{chen2013dynamic}
H.~Chen, H.~A. Rakha, and C.~C. McGhee, ``Dynamic travel time prediction using
  pattern recognition,'' in \emph{20th World Congress on Intelligent
  Transportation Systems}.\hskip 1em plus 0.5em minus 0.4em\relax TU Delft,
  2013.

\bibitem{wang2014travel}
Y.~Wang, Y.~Zheng, and Y.~Xue, ``Travel time estimation of a path using sparse
  trajectories,'' in \emph{SIGKDD}.\hskip 1em plus 0.5em minus 0.4em\relax ACM,
  2014, pp. 25--34.

\bibitem{wang2016simple}
H.~Wang, Y.~H. Kuo, D.~Kifer, and Z.~Li, ``A simple baseline for travel time
  estimation using large-scale trip data,'' in \emph{SIGSPATIAL GIS}.\hskip 1em
  plus 0.5em minus 0.4em\relax Association for Computing Machinery, 2016,
  p.~61.

\bibitem{yuan2011t}
J.~Yuan, Y.~Zheng, X.~Xie, and G.~Sun, ``T-drive: Enhancing driving directions
  with taxi drivers' intelligence,'' \emph{IEEE Transactions on Knowledge and
  Data Engineering}, vol.~25, no.~1, pp. 220--232, 2011.

\bibitem{lecun2015deep}
Y.~LeCun, Y.~Bengio, and G.~Hinton, ``Deep learning,'' \emph{nature}, vol. 521,
  no. 7553, p. 436, 2015.

\bibitem{krizhevsky2012imagenet}
A.~Krizhevsky, I.~Sutskever, and G.~E. Hinton, ``Imagenet classification with
  deep convolutional neural networks,'' in \emph{NeurIPS}, 2012, pp.
  1097--1105.

\bibitem{larochelle2009exploring}
H.~Larochelle, Y.~Bengio, J.~Louradour, and P.~Lamblin, ``Exploring strategies
  for training deep neural networks,'' \emph{Journal of machine learning
  research}, vol.~10, no. Jan, pp. 1--40, 2009.

\bibitem{li2018multi}
Y.~Li, K.~Fu, Z.~Wang, C.~Shahabi, J.~Ye, and Y.~Liu, ``Multi-task
  representation learning for travel time estimation,'' in \emph{SIGKDD}.\hskip
  1em plus 0.5em minus 0.4em\relax ACM, 2018, pp. 1695--1704.

\bibitem{wang2018will}
D.~Wang, J.~Zhang, W.~Cao, J.~Li, and Y.~Zheng, ``When will you arrive?
  estimating travel time based on deep neural networks,'' in \emph{AAAI}, 2018.

\bibitem{fu2019deepist}
T.-y. Fu and W.-C. Lee, ``Deepist: Deep image-based spatio-temporal network for
  travel time estimation,'' in \emph{ACM CIKM}, 2019, pp. 69--78.

\bibitem{lan2019travel}
W.~Lan, Y.~Xu, and B.~Zhao, ``Travel time estimation without road networks: an
  urban morphological layout representation approach,'' in \emph{IJCAI}.\hskip
  1em plus 0.5em minus 0.4em\relax AAAI Press, 2019, pp. 1772--1778.

\bibitem{zhang2018deeptravel}
H.~Zhang, H.~Wu, W.~Sun, and B.~Zheng, ``Deeptravel: a neural network based
  travel time estimation model with auxiliary supervision,'' in \emph{IJCAI},
  2018, pp. 3655--3661.

\bibitem{chopra2005learning}
S.~Chopra, R.~Hadsell, and Y.~LeCun, ``Learning a similarity metric
  discriminatively, with application to face verification,'' in \emph{CVPR},
  vol.~1.\hskip 1em plus 0.5em minus 0.4em\relax IEEE, 2005, pp. 539--546.

\bibitem{schroff2015facenet}
F.~Schroff, D.~Kalenichenko, and J.~Philbin, ``Facenet: A unified embedding for
  face recognition and clustering,'' in \emph{CVPR}, 2015, pp. 815--823.

\bibitem{sohn2016improved}
K.~Sohn, ``Improved deep metric learning with multi-class n-pair loss
  objective,'' in \emph{NeurIPS}, 2016, pp. 1857--1865.

\bibitem{hochreiter1997long}
S.~Hochreiter and J.~Schmidhuber, ``Long short-term memory,'' \emph{Neural
  computation}, vol.~9, no.~8, pp. 1735--1780, 1997.

\bibitem{paszke2019pytorch}
A.~Paszke, S.~Gross, F.~Massa, A.~Lerer, J.~Bradbury, G.~Chanan, T.~Killeen,
  Z.~Lin, N.~Gimelshein, L.~Antiga \emph{et~al.}, ``Pytorch: An imperative
  style, high-performance deep learning library,'' in \emph{NeurIPS}, 2019, pp.
  8024--8035.

\bibitem{kingma2015adam}
D.~Kingma and J.~Ba, ``Adam: A method for stochastic optimization,''
  \emph{ICLR, San Diego}, 2015.

\end{thebibliography}

\end{document}